\documentclass{bvm} 
\usepackage{booktabs} 
\begin{document}

\newcommand{\bvmyear}{2026}

\selectlanguage{english} 

\title{Multimodal classification of Radiation-Induced Contrast Enhancements and tumor recurrence using deep learning}


\titlerunning{BVM \bvmyear}

\author{
    \fname{Robin \lname[0000-0002-6187-3636]{Peretzke} \inst{1,2,3} \affiliation{DKFZ Heidelberg} \authorsEmail{robin.peretzke@dkfz-heidelberg.de}},
    \fname{Marlin \lname{Hanstein} \inst{1} \affiliation{DKFZ Heidelberg} \authorsEmail{marlin.hanstein@dkfz-heidelberg.de}},
    \fname{Maximilian {Fischer} \inst{1,2,4,5} \affiliation{DKFZ Heidelberg} \authorsEmail{maximilian.fischer@dkfz-heidelberg.de}},
    \fname{Lars Badhi {Wessel} \inst{6,7,8,9} \affiliation{Universitätsklinikum Heidelberg / DKFZ Heidelberg} \authorsEmail{lars.wessel@dkfz-heidelberg.de}},
    \fname{Obada {Alhalabi} \inst{10} \affiliation{Universitätsklinikum Heidelberg, Neurochirurgie} \authorsEmail{obada.alhalabi@med.uni-heidelberg.de}},
    \fname{Sebastian {Regnery} \inst{6,7,8,9} \affiliation{Universitätsklinikum Heidelberg / HIT} \authorsEmail{sebastian.regnery@med.uni-heidelberg.de}},
    \fname{Andreas {Kudak} \inst{6,7} \affiliation{Universitätsklinikum Heidelberg / HIRO} \authorsEmail{andreas.kudak@med.uni-heidelberg.de}},
    \fname{Maximilian {Deng} \inst{6,7,8,9} \affiliation{Universitätsklinikum Heidelberg / HIT} \authorsEmail{maximilian.deng@med.uni-heidelberg.de}},
    \fname{Paul Vincent {Naser} \inst{2,10} \affiliation{Universitätsklinikum Heidelberg, Neurochirurgie} \authorsEmail{paul.naser@med.uni-heidelberg.de}},
    \fname{Jan-Oliver {Neumann} \inst{10} \affiliation{Universitätsklinikum Heidelberg, Neurochirurgie} \authorsEmail{oliver.neumann@med.uni-heidelberg.de}},
    \fname{Tanja {Eichkorn} \inst{6,7,8,9} \affiliation{Universitätsklinikum Heidelberg / NCT Heidelberg} \authorsEmail{tanja.eichkorn@med.uni-heidelberg.de}},
    \fname{Philipp {Hoegen-Saßmannshausen} \inst{6,7,8,9} \affiliation{Universitätsklinikum Heidelberg / HIT} \authorsEmail{philipp.hoegen@med.uni-heidelberg.de}},
    \fname{Fabian {Allmendinger} \inst{6,7,8} \affiliation{Universitätsklinikum Heidelberg} \authorsEmail{fabian.allmendinger@med.uni-heidelberg.de}},
    \fname{Jan-Hendrik {Bolten} \inst{6,7,8} \affiliation{Universitätsklinikum Heidelberg} \authorsEmail{jan-hendrik.bolten@med.uni-heidelberg.de}},
    \fname{Philipp {Schröter} \inst{6,7,8} \affiliation{Universitätsklinikum Heidelberg} \authorsEmail{philipp.schroeter@med.uni-heidelberg.de}},
    \fname{Christine {Jungk} \inst{10} \affiliation{Universitätsklinikum Heidelberg, Neurochirurgie} \authorsEmail{christine.jungk@med.uni-heidelberg.de}},
    \fname{Jürgen Peter {Debus} \inst{1,6,7,8,9,11} \affiliation{Universitätsklinikum Heidelberg / DKFZ Heidelberg} \authorsEmail{juergen.debus@med.uni-heidelberg.de}},
    \fname{Peter {Neher} \inst{1,6} \affiliation{DKFZ Heidelberg} \authorsEmail{peter.neher@dkfz-heidelberg.de}},
    \fname{Laila {König} \inst{6,7,8,9} \affiliation{Universitätsklinikum Heidelberg / HIT} \authorsEmail{laila.koenig@med.uni-heidelberg.de}},
    \fname{Klaus {Maier-Hein} \inst{1,2,3} \affiliation{DKFZ Heidelberg} \authorsEmail{k.maier-hein@dkfz-heidelberg.de}}
}

\authorrunning{Peretzke, Hanstein, Fischer et al.}

\institute{
\inst{1} Deutsches Krebsforschungszentrum (DKFZ), Division of Medical Image Computing, Heidelberg, Germany\\
\inst{2} Medical Faculty, Heidelberg University, Heidelberg, Germany\\
\inst{3} Pattern Analysis and Learning Group, Department of Radiation Oncology, Heidelberg University Hospital, Heidelberg, Germany\\
\inst{4} Research Campus M\textsuperscript{2}OLIE, Mannheim, Germany\\
\inst{5} German Cancer Consortium (DKTK), DKFZ Core Center Heidelberg, Heidelberg, Germany\\
\inst{6} Universitätsklinikum Heidelberg, Heidelberg, Germany\\
\inst{7} Heidelberger Institut für Radioonkologie (HIRO), Heidelberg, Germany\\
\inst{8} National Center for Tumor Diseases (NCT), NCT Heidelberg, a partnership between DKFZ and University Medical Center Heidelberg\\
\inst{9} Heidelberger Ionenstrahl-Therapiezentrum (HIT), Heidelberg, Germany\\
\inst{10} Universitätsklinikum Heidelberg, Neurochirurgie, Heidelberg, Germany\\
\inst{11} Deutsches Krebsforschungszentrum (DKFZ), Clinical Cooperation Unit Radiation Oncology, Heidelberg, Germany\\
}

\email{maximilian.fischer@dkfz-heidelberg.de}

\maketitle

\begin{abstract} The differentiation between tumor recurrence and radiation-induced contrast enhancements in post-treatment glioblastoma patients remains a major clinical challenge. Existing approaches rely on clinically sparsely available diffusion MRI or do not consider radiation maps, which are gaining increasing interest in the tumor board for this differentiation. We introduce \textbf{RICE-NET}, a multimodal 3D deep learning model that integrates longitudinal MRI data with radiotherapy dose distributions for automated lesion classification, using conventional T1-weighted MRI data. Using a cohort of 92 patients, the model achieved a performance of 0.92 [F1] on an independent test set. During extensive ablation experiments, we quantified the contribution of each timepoint and modality and showed that reliable classification largely depends on the radiation map. Occlusion-based interpretability analyses further confirmed the model’s focus on clinically relevant regions. These findings highlight the potential of multimodal deep learning to enhance diagnostic accuracy and support clinical decision-making in neuro-oncology.
\end{abstract}

\section{Introduction}
Medicine has always been an exercise in balance between benefit and harm, precision and uncertainty \cite{haanen2022management}. This tension is especially apparent in the treatment of brain tumors, where radiation, a life-saving therapy, also carries the potential to injure the very tissue it seeks to protect. While radiotherapy is indispensable in eliminating residual tumor cells after surgical resection, its necrotic effect is not exclusive to malignant tissue. As a result, new or progressive contrast-enhancing lesions that appear during follow-up imaging often pose a crucial diagnostic dilemma: do they represent tumor recurrence or Radiation Induced Contrast Enhancements (RICE)? Both malignancies are expressed via similar appearing contrast-enhanced regions in MR imaging. Although recent studies indicate a higher incidence of RICE in recent years, the differentiation between tumor recurrence and RICE remains challenging \cite{eichkorn2024radiation,eichkorn2023analysis}. Today, the clinical workflow for distinguishing between RICE and tumor recurrence is based on a complex and time-consuming process within an interdisciplinary tumor board. This process involves re-evaluation of the imaging trajectory, consisting of pre- and post-operative imaging, various follow-up scans around radiation therapy, as well as the contrast event MRI scan, which is generated to verify new symptoms that indicate a novel lesion that appears contrast-enhanced.

In this work, we present RICE-NET, a multimodal deep learning approach for differentiating between radiation-induced changes and glioblastoma multiforme (GBM) tumor recurrence. The network integrates longitudinal MRI data as well as radiotherapy plans to enable automated and early classification of post-treatment contrast-enhancing lesions. While several recent studies have reported promising results using advanced diffusion imaging for lesion classification \cite{bernhardt2022degro,wang2016differentiating}, these approaches have not yet been applied to conventional MRI sequences that are more widely available in clinical practice. Moreover, most existing studies do not incorporate dosimetric data and neglect the longitudinal evolution of imaging findings.

We systematically evaluate the individual contributions of each input modality to the overall classification performance of RICE-NET, providing insights into their relative diagnostic value. We show that radiation dosage (RD) information plays a critical role in achieving accurate classification. This contribution highlights the importance of incorporating radiation dose distributions into the clinical decision-making process.

\section{Materials and Methods}
This study utilized multimodal, longitudinal data to develop a framework for differentiating RICE from tumor recurrence in patients with GBM. This section describes the dataset, model architecture, experimental setup, and evaluation procedures.

\subsection{Dataset Description}
RICE-Net was developed on a patient cohort, consisting of 92 GBM patients who received treatment at the University Clinic Heidelberg (UKHD), with the approval of the institutional ethics committee. For the development and evaluation, this cohort was divided into a training and validation set of 80 patients (comprising 48 with tumor recurrence and 32 with Radiation-Induced Contrast Enhancement (RICE)) and a separate, independent test set of 12 patients (7 with tumor recurrence and 5 with RICE) with a separation subject level.
Within the dataset, for each subject three different imaging volumes are available. First, a post-operative, T1-weighted contrast-enhanced MRI, which is referred to as "MRI post-OP". This scan was acquired after the initial surgical resection and serves as a baseline image to plan the subsequent radiation therapy. Second, a follow-up T1-weighted contrast-enhanced MRI, termed the "MRI event". This scan was captured at the critical time of diagnostic uncertainty when new contrast-enhancing lesions were first detected, representing either true tumor recurrence or RICE. The indication for this scan was either a regular follow-up scan or the clarification of unclear symptoms. Third, the "RD map", a 3D map detailing the spatial distribution and cumulative dose of radiation delivered during the patient's radiotherapy treatment.
The ground truth labels (“tumor recurrence” or “RICE”) were confirmed and provided by clinical partners at UKHD based on biopsy results.

\begin{figure}[H]
    \centering
    \includegraphics[width=0.95\linewidth]{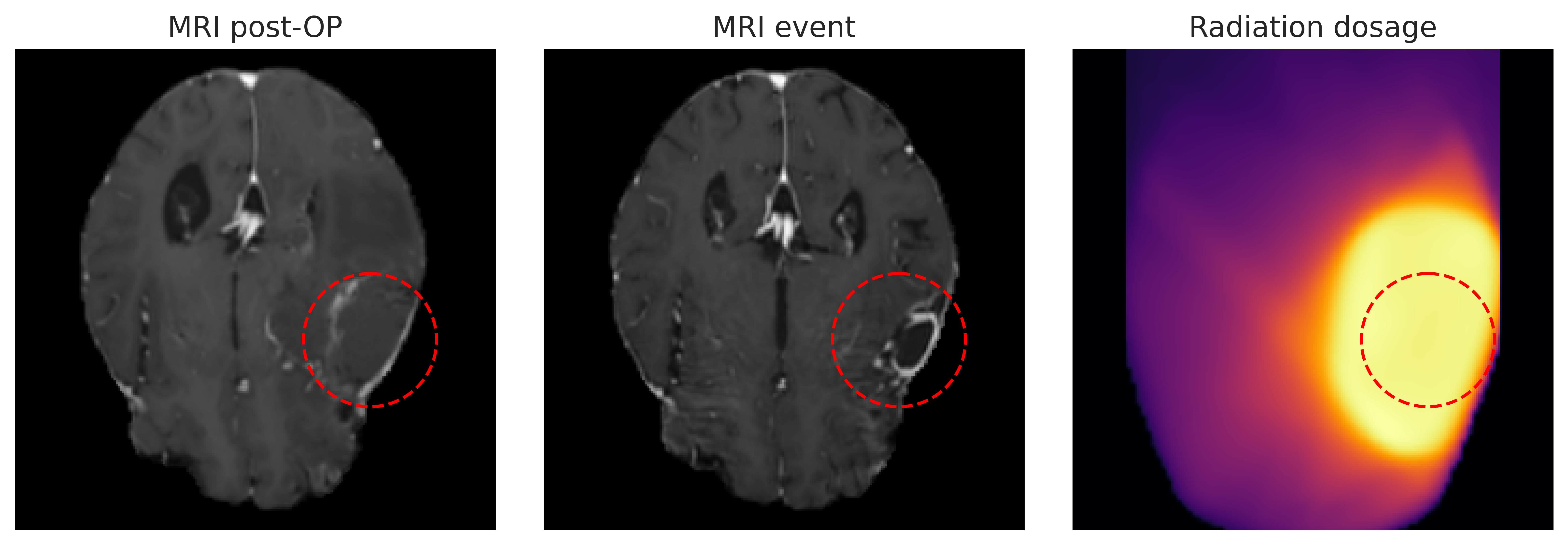}
    \caption{Axial slice of sample subject with tumor recurrence. On the left, the post-operative MRI with the resection area highlighted with a red circle. The middle image shows the new progression, which is to be classified as recurrence or RICE. On the right, the radiation treatment plan is displayed, with the isocenter positioned in the region of the resection cavity.}
    \label{fig:samplesubject}
\end{figure}

In Figure \ref{fig:samplesubject} the MRI scan after resection and at the moment of the new contrast event, and the radiotherapy plan for a sample subject are visualized.  

All imaging volumes underwent a standardized preprocessing pipeline. The volumes were resampled to an isometric voxel spacing and co-registered using the Advanced Normalization Tools (ANTS) \cite{tustison2021antsx}. The brain was extracted from the skull using HD-BET \cite{isensee2019automated}, and signal intensities were normalized via z-scoring. Finally, the volumes were cropped to a consistent size of 224x224x224 voxels centered on the brain. For those patients, whom only a single fraction of the radiation treatment was available, the total RD was calculated by scaling a single-fraction dose map with the number of fractions, assuming consistent radiation delivery.

\subsection{Deep Learning Architecture}
A 3D residual network (ResNet18) architecture, adapted for volumetric medical image classification, was implemented using the MONAI\footnote{\url{https://monai.io/}} framework \cite{cardoso2022monai}. The network extends the original 2D ResNet design \cite{he2016deep} to three dimensions, enabling the extraction of spatially coherent features across MRI volumes. Residual connections facilitate efficient gradient flow, improving convergence stability and representational capacity, which are both critical for medical image analysis tasks.

The model comprises an initial 3D convolutional layer followed by four residual blocks with batch normalization and ReLU activations, concluding with global average pooling and a fully connected classification layer. ResNet18 was chosen for its balance between expressiveness and computational efficiency, reducing overfitting risks while maintaining strong performance on limited clinical datasets.

\subsection{Experimental Setup}
To systematically assess the diagnostic contribution of each data modality, we conducted a series of ablation experiments while keeping the network architecture fixed. In these experiments, we varied the input modalities and their combinations. The data folds were fixed across all experiments and were defined at the subject level. The following input datapoints were considered: (1) MRI post-OP, (2) MRI event, (3) RD followed by various combinations of the individual data points: (4) MRI post-OP + MRI event, (5) MRI post-OP + RD, (6) MRI event + RD, and (7) all available data points combined. For each different scenario, an individual model was trained.

For experiments involving multiple modalities, the inputs were concatenated along the channel dimension. All models were trained for 800 epochs using a five-fold cross-validation on the 80-patient training set. For training, the Adam optimizer and cross-entropy loss are implemented, along with a weighted random sampler to ensure equal representation of both classes during training. As augmentations, we used elastic deformations, rotations, scaling, Gaussian noise, and brightness and gamma adjustments. As primary evaluation metric, we selected the macro F1-score. This metric is the unweighted average of the F1-scores computed for each class (RICE and tumor recurrence). The F1-score, the harmonic mean of precision and recall, provides a balanced measure that remains robust under class imbalance.

For model interpretability, occlusion sensitivity maps are evaluated to identify critical input regions by systematically masking image parts and observing the effect on the output \cite{zeiler2014visualizing}. We occluded small cubic 3D regions synchronously across all co-registered volumes and measured the resulting change in output probability, highlighting regions most influential for distinguishing RICE from tumor recurrence. The maps provide an intuitive view of model attention that aids clinical interpretation.

\section{Results}

\begin{figure}[h]
    \centering
    \includegraphics[width=.9\linewidth]{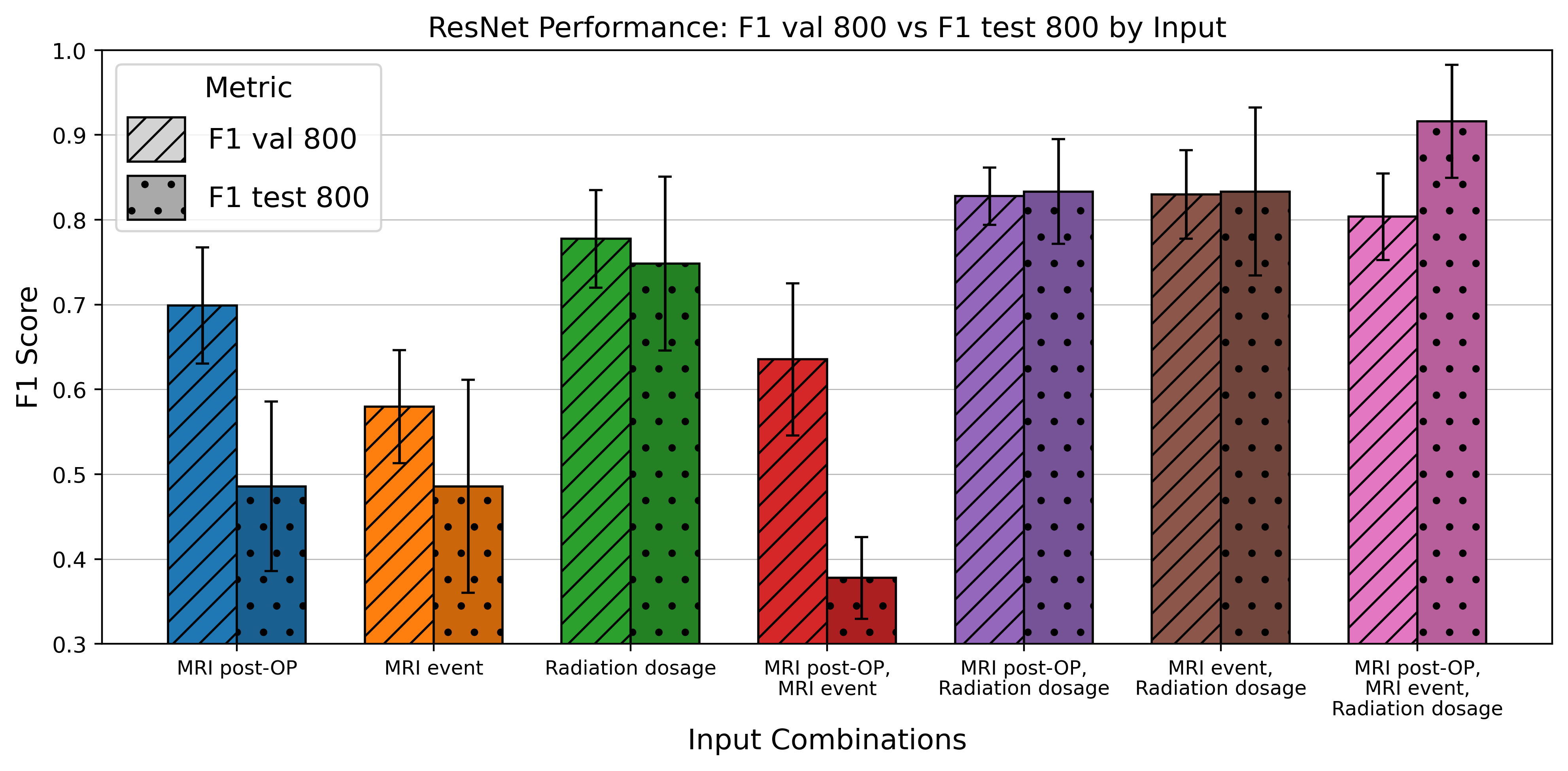}
    \caption{F1 Macro after 800 training epochs on validation data (striped) aggregated across all folds and majority vote on the test data (dotted) by input volume combinations with cross validation standard deviation as error bars.}
    \label{fig:resnet_val_vs_test_f1}
\end{figure}

The performance of the model architecture on our dataset was evaluated with a series of ablation studies designed to assess the relative contribution of each input modality. The results of these experiments, summarized in Figure~\ref{fig:resnet_val_vs_test_f1}, report model performance in terms of the F1-score. Metrics were obtained from five-fold cross-validation and from majority voting on an independent hold-out test set to evaluate generalization performance. In the single-modality experiments, the radiation dosage map served as the most informative input, resulting in a macro F1-score of 0.78 on the validation set. The post-operative MRI and contrast event MRI result in lower F1 scores of 0.70 and 0.58, respectively. Performance was evaluated across multiple modality combinations. Integrating MRI with the radiation dose (RD) map yielded the best validation results, with the contrast-event MRI + RD model reaching an F1-score of 0.83 and the post-operative MRI + RD model achieving 0.828. Using all three inputs produced a validation F1-score of 0.804. On the independent test cohort, an ensemble of cross-validated models achieved an F1-score of 0.916.

Figure \ref{fig:resnet_val_vs_test_f1} shows an example occlusion map for a correctly classified RICE case, highlighting the spatial regions within the input volumes that most strongly contributed to the model’s prediction.

\begin{figure}[h]
    \centering
    \includegraphics[width=1\linewidth]{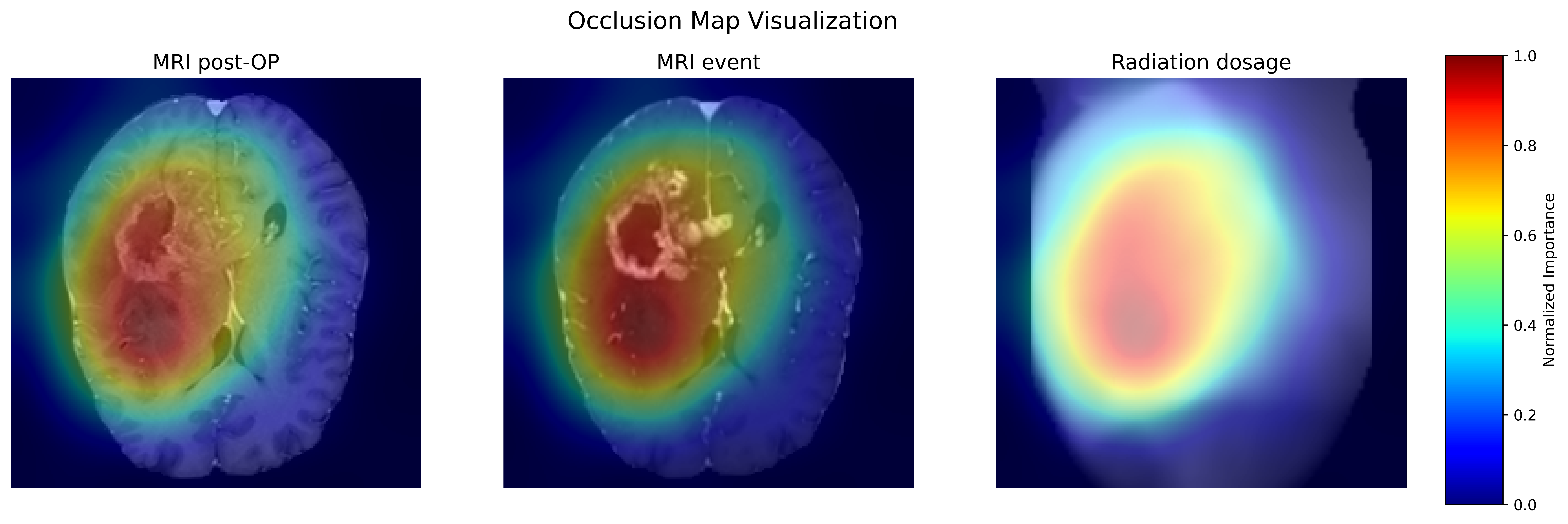}
    \caption{Occlusion visualization with RICE overlaid over all three inputs.}
    \label{fig:occlusion_RICE}
\end{figure}


\section{Discussion}
We present RICE-Net, a multichannel 3D ResNet-18 for differentiating RICE from tumor recurrence using MRI and radiotherapy data. The model was trained on post-operative and contrast-event MRI volumes together with radiation dose distributions, and ablation studies quantified the contribution of each modality. Models using only radiation maps already performed well, indicating that spatial dose distribution provides strong predictive information on tissue response. Combining MRI with dose maps further improved results, with the best performance achieved when all modalities were integrated, confirming their complementary value. Notably, combining the radiation plan with event MRI did not outperform the combination with post-operative MRI, reflecting the clinical challenge of distinguishing true progression from treatment effects using MRI alone. Post-operative MRI and dose information may already encode early markers of RICE risk, enabling earlier prediction. A pronounced performance gap between cross-validation and test F1-scores, especially in MRI-only experiments ($\approx$ 0.35 lower on test), highlights statistical uncertainty from limited cohort size. Occlusion maps showed strong correlation with high-dose regions, confirming reliance on dosimetric information while also emphasizing contrast-enhancing lesions, indicating multimodal reasoning.

The main limitations are the small cohort size, lack of unaffected subjects, and the simple channel-wise fusion, which may miss complex MRI–dose interactions. Future work should expand the dataset, include non-recurrent cases, and explore advanced fusion and clinical variable integration. Multi-institutional validation will be key to assessing RICE-Net’s potential for early post-treatment RICE prediction.



\printbibliography
\end{document}